\documentclass{INTERSPEECH2023}

\interspeechcameraready

\usepackage{bm}
\usepackage{multirow}
\usepackage{subfigure}
\usepackage{tikz}
\usetikzlibrary{bayesnet}
\usetikzlibrary{arrows}
\usepackage{color}
\usetikzlibrary{backgrounds}

\def\rvx{{\mathbf{x}}}
\def\rvy{{\mathbf{y}}}
\def\rvz{{\mathbf{z}}}

\DeclareMathOperator*{\argmax}{arg\,max}

\newcommand*{\su}[1]{^{\mkern-1.2mu #1}}    
\newcommand{\en}{\textrm{en}}
\newcommand{\fr}{\textrm{fr}}
\newcommand{\pt}{\textrm{pt}}

\newcommand{\taq}{\textrm{taq}}
\newcommand{\chrf}{\textsc{chrF2}}

\title{Strategies for improving low resource speech to text translation relying on pre-trained ASR models}
\name{Santosh Kesiraju$^1$, Marek Sarva\v{s}$^1$, Tom\'{a}\v{s} 
Pavli\v{c}ek$^2$, C\'{e}cile Macaire$^3$, Alejandro Ciuba$^4$}
\address{
  $^1$Speech@FIT, Brno University of Technology, Czechia. $^2$Phonexia, 
  Czechia. \\
  $^3$Univ. Grenoble Alpes, France. $^4$University of Pittsburgh, USA.}
\email{kesiraju@fit.vutbr.cz, xsarva00@stud.fit.vutbr.cz, 
tomas.pavlicek@phonexia.com, cecile.macaire@univ-grenoble-alpes.fr, 
alejandrociuba@pitt.edu}

\begin{document}

\maketitle

\begin{abstract}


  This paper presents techniques and findings for improving the performance of low-resource speech to text translation (ST). We conducted experiments on both simulated and real-low resource setups, on language pairs English - Portuguese, and Tamasheq - French respectively. Using the encoder-decoder framework for ST, our results show that a multilingual automatic speech recognition system acts as a good initialization under low-resource scenarios. Furthermore, using the CTC as an additional objective for translation during training and decoding helps to reorder the internal representations and improves the final translation. Through our experiments, we try to identify various factors (initializations, objectives, and hyper-parameters) that contribute the most for improvements in low-resource setups. With only 300 hours of pre-training data, our model achieved 7.3 BLEU score on Tamasheq - French data, outperforming prior published works from IWSLT 2022 by 1.6 points.
\end{abstract}

\section{Introduction}

Speech translation (ST) systems consume speech (features) from source language as input and generate text in the target language. A cascaded approach to this task involves passing speech through an automatic speech recognition (ASR) system that generates (decodes) $n$-best discrete text-hypotheses in source language, which are then passed on to a text-based machine translation (MT) system to generate the text in target language (Fig.~\ref{fig:block}a). Here, the errors from the ASR outputs are \textit{likely} to be propagated to the MT system. End-to-end approaches aim to overcome such errors by establishing a continuous (differentiable) path from input source speech to target translations (Fig.~\ref{fig:block}b)~\cite{Hari:2020:Joint,dalmia-etal-2021-searchable}.
End-to-end approaches based on encoder-decoder architectures also make use of source transcriptions to provide additional supervision (Fig.~\ref{fig:block}c)~\cite{inaguma-etal-2020-espnet}. There were also attempts to train a direct speech translation system without relying on source text, however such approaches were studied only on high-resource scenarios (Fig.~\ref{fig:block}d)~~\cite{ZhangHS22}. For high resource scenarios, the ASR on source language can be trained on huge amounts of available transcribed data, and the MT can be also trained on massive parallel data. Such trained modules can be used as initializations in any of the above frameworks.

\begin{figure}[t]
  \centering
  \begin{subfigure}{
  \begin{tikzpicture}
    [
    rect/.style={minimum size=0.6cm,text width=1cm, align=center,rectangle,draw,rounded corners},
    var/.style={minimum size=1pt,circle},
    arr/.style={->,>=stealth',semithick},
    ]
    \node (x) [var] at (-1.4, 0) {$\mathbf{x}$};
    \node (asr) [rect] at (0,0) {ASR};
    \node (y) [var] at (1.4, 0) {$\mathbf{y}$} ;
    \node (mt) [rect] at (3,0) {MT};
    \node (z) [var] at (4.4,0) {$\mathbf{z}$};
    \path
    (x)   edge [arr] (asr)
    (asr) edge [arr] (y)
    (y)   edge [arr] (mt)
    (mt)  edge [arr] (z)
    ;
  \end{tikzpicture}
  }
  \caption*{a. Cascade system}
  \end{subfigure}
  \begin{subfigure}{
  \begin{tikzpicture}
  [
  rect/.style={minimum size=0.6cm,text width=1cm, align=center, rectangle,draw,rounded corners},
  var/.style={minimum size=1pt,circle},
  arr/.style={->,>=stealth',semithick},
  ]
  \node (x) [var] at (-1.4, 0) {$\mathbf{x}$};
  \node (asr) [rect] at (0, 0) {ASR};
  \node (y) [var] at (1.4, 0.5) {$\mathbf{y}$} ;
  \node (h) [var] at (1.5, -0.2) {$\mathbf{h}$} ;
  \node (mt) [rect] at (3, 0) {MT};
  \node (z) [var] at (4.4, 0) {$\mathbf{z}$};
  \draw
  (x)   edge [arr] (asr)
  (asr) edge [arr, midway,bend right=20] (y)
  (asr)   edge [arr] (mt)
  (mt)  edge [arr] (z)
  ;
    \end{tikzpicture}
  }
  \caption*{b. Joint training with end-to-end differentiability}
  \end{subfigure}
  \begin{subfigure}{
  \begin{tikzpicture}
  [
  rect/.style={minimum size=0.6cm,text width=1cm, align=center, rectangle,draw,rounded corners},
  var/.style={minimum size=1pt,circle},
  arr/.style={->,>=stealth',semithick},
  ]
  \node (x) [var] at (-1.4, 0) {$\mathbf{x}$};
  \node (asr) [rect] at (0, 0) {Encoder};
  \node (y) [var] at (1.4, 0.5) {$\mathbf{y}$} ;
  \node (mt) [rect] at (3, 0) {Decoder};
  \node (z) [var] at (4.4, 0) {$\mathbf{z}$};
  \draw
  (x)   edge [arr] (asr)
  (asr) edge [arr, midway,bend right=20] (y)
  (asr)   edge [arr] (mt)
  (mt)  edge [arr] (z)
  ;
  \end{tikzpicture}
  }
  \caption*{c. End-to-end model with transcriptions as auxiliary objective.}
  \end{subfigure}
  \begin{subfigure}{
  \begin{tikzpicture}
  [
  rect/.style={minimum size=0.6cm,text width=1cm, align=center,rectangle,draw,rounded corners},
  var/.style={minimum size=1pt,circle},
  arr/.style={->,>=stealth',semithick},
  ]
  \node (x) [var] at (-1.4, 0) {$\mathbf{x}$};
  \node (asr) [rect, text width=1.1cm] at (0, 0) {Encoder};
  \node (y) [var] at (1.4, 0.5) {$\mathbf{z}$} ;
  \node (mt) [rect, text width=1.1cm] at (3, 0) {Decoder};
  \node (z) [var] at (4.4, 0) {$\mathbf{z}$};
  \draw
  (x)   edge [arr] (asr)
  (asr) edge [arr, midway,bend right=20] (y)
  (asr)   edge [arr] (mt)
  (mt)  edge [arr] (z)
  ;
  \end{tikzpicture}
  }
  \caption*{d. End-to-end model with translations as auxiliary objective.}
  \end{subfigure}
  \caption{Cascaded and end-to-end frameworks for speech translation. $\mathbf{x}$ is the input speech (features), $\mathbf{y}$ is the corresponding text transcriptions, and $\mathbf{z}$ is the target text translations. $\mathbf{h}$ is the hidden representation from ASR that establishes the continuous path between ASR and MT models. The ASR, MT, encoder and decoder modules can be initialized from various kinds of pre-trained models.}
  \label{fig:block}
  \vspace{-0.5cm}
\end{figure}

\begin{table}
  \centering
  \caption{Initialization options for encoder-decoder based speech translations systems.}
  \label{tab:init_options}
  \begin{tabular}{llr}
    \toprule
    \text{Encoder init.}   & \text{Decoder init.} & \text{Aligned?} \\
    \midrule
    Encoder from ASR   & Decoder from ASR    & Yes  \\
    Encoder from ASR   & Decoder from MT     & No \\
    Encoder from SSL   & Decoder from MT     & No \\
    Random             & Random              & No \\
    \bottomrule
  \end{tabular}
\vspace{-0.1cm}
\end{table}

However, such a luxury is not available in low-resource scenarios, where neither source speech transcriptions, nor source to target parallel text data are available. Moreover, the amount of speech translation training data can also be very limited (e.g. $<20$ hours), which is also the scenario for most of the experiments and analyses in this paper. Automatic translation of speech from a low-resource to high-resource language has applications in topic detection~\cite{strassel-tracey-2016-lorelei,BansalKLG20}.
 In such low-resource scenarios, one can rely on transfer learning, where the ST model or parts of it are either initialized from a \textit{target-language} ASR or MT or a speech representation model based on self-supervised learning (SSL). More specifically, in an encoder-decoder framework for speech translation, the speech encoder can be initialized from a pre-trained ASR~\cite{BansalKLLG19} or SSL~\cite{xls-r:2022}, whereas the decoder can be initialized from a pre-trained ASR~\cite{StoianBG20} or MT~\cite{mbart:2020} model. The model can then be fine-tuned using the target speech translation data. Depending on the choice of initializations, the encoder and decoder can either be aligned or misaligned, i.e., the contextual representations from encoder live in a subspace different than that of the representations in the decoder. Moreover, the vocabulary of an ASR and MT system can differ, which also contributes to the misalignment. Table~\ref{tab:init_options} summarizes the various initialization options and the consequent alignments. The benefit of initializations from large pre-trained models is diminished when the fine-tuning data is very low, which can be attributed to the misaligned representations during initialization.

Such a problem of misaligned initialization doesn't arise when both the encoder and decoder are initialized from a pre-trained ASR. However, the ASR models assume monotonic alignment between the input speech and target text, which is not true in the case of speech translation. Here the challenge is to learn the re-ordering with limited amount of ST training data. While there are numerous approaches and analysis on high-resource speech translation~\cite{chuang-etal-2021-investigating,bentivogli-etal-2021-cascade,joint_vs_cascade_2022,yan:2023:eacl}, there is scope for studying these techniques in low-resource scenarios.

%

%
%

\subsection{Related works}
\label{ssec:related}
Prior works~\cite{BansalKLLG19,StoianBG20} have shown that a speech translation system initialized from a monolingual ASR built on target language could benefit in low-resource speech translation. The authors concluded that pre-training on any language could still yield a benefit, however the use of pre-trained multilingual ASR is not fully explored in their work.

The Connectionist-temporal classification (CTC)~\cite{Graves:2006:CTC} was originally proposed for ASR. The CTC model built on RNN encoder assume a monotonic alignment between the input speech (features) to the target tokens, which is not suitable for speech translation. Chuang et al.~\cite{chuang-etal-2021-investigating} have shown that transformers trained with CTC objective for speech translation can learn to reorder. This has motivated other works exploring direct speech translation with CTC as an auxiliary objective only during training~\cite{ZhangHS22}. More recently, Yan et al.~\cite{yan-etal-2022-cmus,yan:2023:eacl} have seen the benefits of joint training and decoding for speech translation. However, their models and experiments were mostly focused to mid-to-high resource language where source transcriptions are also available.

In the recent findings from IWSLT 2022 low-resource track for Tamasheq $\rightarrow$ French speech translation task, the majority of the techniques involving large multilingual SSL models (XLS-R) and pre-trained MT models (mBART) have shown very poor results~\cite{iwslt-2022-findings,boito-etal-2022-speech}. This motivated us to revisit the strategies for training low-resource speech translation.
\subsection{Contributions of the paper}
\begin{itemize}
  \item Study of pre-trained multilingual ASR as initialization for low-resource speech translation with joint training and decoding with CTC objective in low-resource setups.
  \item Extensive analysis on the effect of various initialization, auxiliary objectives, hyperparameters and amounts of fine-tuning data, identifies the most important factors that contribute most to the improvements.
  \item On low-resource Tamasheq $\rightarrow$ French task, our ST model initialized from a pre-trained multilingual ASR with only 300 hours training data achieved 7.3 BLEU score, which is +1.6 points higher than the best published result from IWSLT'22.
\end{itemize}
\begin{table}[!t]
  \centering
  \setlength{\tabcolsep}{0.52em}
  \caption{Statistics of speech translation data.}
  \label{tab:st_data_stats}
  \begin{tabular}{crl|rl|rl}
    \toprule
    & \multicolumn{6}{c}{\text{Speech translation data: hours (utterances)}} \\
    \text{Direction}  & \multicolumn{2}{c|}{\text{Training}} & \multicolumn{2}{c|}{\text{Dev.}} & \multicolumn{2}{c}{\text{Test}}   \\
    \midrule
    \taq $\rightarrow$ \fr   & 13.8    & (4444)   & 1.9   & (581)  & 2.0 & (804)  \\
    \en  $\rightarrow$ \pt   & 292.5   & (184.3k)  & 3.2  & (2022) & 3.7 & (2305) \\
    \midrule
    \multicolumn{5}{l}{\text{Low-resource simulation splits}} &  &    \\
    \midrule
    \en  $\rightarrow$ \pt & 50.0  & (31.5k) & 3.2 & (2022) & 3.7 & (2305) \\
    \en  $\rightarrow$ \pt & 16.4  & (10.5k) & 3.2 & (2022) & 3.7 & (2305) \\
    \bottomrule
  \end{tabular}
\end{table}
\section{Methodology}
This section formally introduces the necessary terminology and describes the methods we followed to train ASR and ST systems. The ASR is trained on several examples of paired speech and text $(\rvx\su{s}, \rvy\su{s})$ from one or more (\textit{seen}) languages $s \in \mathcal{S}$. The speech translation systems are trained on pairs $(\rvx\su{u}, \rvz\su{s})$, where the input speech $\rvx\su{u}$ is from an \textit{unseen} language $u \notin \mathcal{S}$, and the target translation text $\rvz\su{s}$ is from \textit{seen} languages $s \in \mathcal{S}$.
\subsection{Training ASR}
A transformer~\cite{Vaswani:NIPS2017} based encoder-decoder architecture with additional CTC layer is used to train the ASR models. For multilingual ASR, we keep a separate vocabulary for each language, which results in a language-specific CTC layer at the output of the encoder, and language-specific input (embedding) and output layers in the decoder. Such an architecture allows us to decode tokens only in the desired target language. The models are trained with joint CTC and attention objective function~\cite{karita19_interspeech}
\begin{equation}
  \mathcal{L}_{\mathrm{asr}}(\rvx\su{s}, \rvy\su{s}) = \lambda \, \mathcal{L}_{\mathrm{ctc}}(\rvx\su{s}, \rvy\su{s}) + (1-\lambda) \, \mathcal{L}_{\mathrm{att}}(\rvx\su{s}, \rvy\su{s}).
\end{equation}
\subsection{Training ST}
The ST models are also based on transformer encoder-decoder architecture and are identical to the ASR models, which allows us to initialize ST models with any pre-trained ASR. More specifically, we are given speech $\rvx\su{u}$ from a previously unseen language $u \notin \mathcal{S}$, and its translation $\rvz\su{s}$ from a language that was already seen, $s \in \mathcal{S}$. The ST model is also trained with joint objective function
\begin{equation}
  \mathcal{L}_{\mathrm{st}}(\rvx\su{u}, \rvz\su{s}) = \alpha \, \mathcal{L}_{\mathrm{ctc}}(\rvx\su{u}, \rvz\su{s}) + (1-\alpha) \, \mathcal{L}_{\mathrm{att}}(\rvx\su{u}, \rvz\su{s}).
\end{equation}
\subsection{Decoding}
A beam search based joint decoding~\cite{karita19_interspeech} that relies on the weighted average of log-likelihoods from both the CTC and transformer decoder modules is used, that produces the most likely hypotheses according to
\begin{equation}
\hat{\rvz} = \argmax_{\rvz} \, \beta \, \log p_{\mathrm{ctc}} (\rvz \mid \rvx) + (1-\beta) \,  \log p_{\mathrm{att}} (\rvz \mid \rvx).
 \end{equation}
%
%
\section{Experimental setup}
\label{sec:exp}
The ST experiments were conducted on two datasets: (i) Tamasheq (\taq) 
$\rightarrow$ French (\fr) from IWSLT'22 evaluation 
campaign~\cite{iwslt-2022-findings,boito-etal-2022-speech}, and (ii) English 
(\en) $\rightarrow$ Portuguese (\pt) from HOW2 dataset~\cite{how2:2018}. The 
latter dataset is mainly used for simulating low-resource setups with various 
amounts of fine-tuning data. Moreover, it also allows is to compare the 
performance against a typical end-to-end system exploiting source transcripts 
and source-target parallel data.
Table~\ref{tab:st_data_stats} presents the ST data statistics, where the bottom half indicates the low-resource simulation splits derived from HOW2 dataset.
\begin{table}[!t]
  \centering
  \setlength{\tabcolsep}{4pt}
  \caption{Statistics of data for training ASR models.}
  \label{tab:asr_data_stats}
  \begin{tabular}{lrrr}
    \toprule
    & \multicolumn{3}{c}{\text{ASR data: hours}} \\
    \text{Languages}& \text{Training} & \text{Dev.} & \text{Test} \\
    \midrule
    \fr     & \{50$\ldots$ 764\}  & 25.5 & 26.1 \\
    \pt & 50                     & 10.3 & 11.1 \\
    de, es, fr, it, pl, pt (6L) & 300 & 124.1 & \{26.1, 11.1\} \\
    \bottomrule
  \end{tabular}
\vspace{-0.1cm}
\end{table}
To train multilingual ASR models, we picked a subset of 6 languages (6L) from Mozilla Common Voice v8.0, including French and Portuguese. We sampled 50 hours of transcribed data for each language, which resulted in 300 hours of training data. For monolingual ASR training, we considered the same 50hr for Portuguese. In case of French, we trained several monolingual ASR systems on various amounts of data: \{50, 100, 200, 300, 764\} hours.
The Table~\ref{tab:asr_data_stats} presents the statistics of data used for ASR training.
\begin{table}[!t]
  \centering
  \caption{Performance of various ASR systems in terms of word (WER) and character error rates (CER).}
  \label{tab:asr_res}
  \begin{tabular}{ccrr|rr}
    \toprule
    & \text{Training} & \multicolumn{2}{c|}{\text{Dev}} & \multicolumn{2}{c}{\text{Test}} \\
    \text{ASR}    & \text{data (in hrs.)} & \text{WER}  & \text{CER}   & \text{WER} & \text{CER} \\
    \midrule
    & & \multicolumn{4}{c}{French (fr)} \\
    \cmidrule{3-6}
    \multirow{5}{*}{Mono (fr)} & 50      & 39.1 & 21.2 & 42.7 & 23.9    \\
    & 100     & 30.3 & 15.4 & 33.9 & 17.9    \\
    & 200     & 23.8 & 11.8 & 27.4 & 14.1    \\
    & 300     & 21.5 & 10.6 & 24.7 & 12.6    \\
    & 764     & 16.8 & 8.1  & 19.8 & 9.9     \\
    Multi (6L)              & 300  & 33.0 & 16.8 & 36.4 & 19.2     \\
    \midrule
    & & \multicolumn{4}{c}{Portuguese (pt)} \\
    \cmidrule{3-6}
    Mono (pt)                & 50      & 27.0 & 11.2 & 29.6 & 12.6 \\
    Multi (6L)               & 300     & 23.3 & 9.1  & 24.7 & 9.8 \\
    \bottomrule
  \end{tabular}
    \vspace{-0.25cm}
\end{table}
\subsection{Model configuration and hyper-parameters}
The input to the model is
$80$-dimensional filter-bank features appended with $3$-dimensional pitch features extracted from the speech signal for every 25ms, with a frame shift of 10ms. The NN model is based on standard transformer encoder-decoder architecture starting with $\mathrm{Conv2d}$ layer with $256$ output channels, kernel size $(3, 3)$, stride $2$. This is followed by $12$ transformer layers in the encoder and $6$ in  the decoder, with $d_{\mathrm{model}}=256$, $d_{\mathrm{ff}}=2048$, $\mathrm{heads}=4$, $\mathrm{dropout}_{\mathrm{ff}}=\{0.1, 0.2, 0.3\}$, $\mathrm{dropout}_{\mathrm{att}}=\{0.0, 0.1\}$. The models are trained for $\{100, 200\}$ epochs with $25000$ warm-up steps and a peak learning rate from $\{5e-3, 1e-2\}$, using \textsc{adam} optimizer. The batch size is varied among, $\{64, 96, 128\}$ depending on the available GPU memory.

The CTC weight $\lambda$ when training ASR models was chosen from 
$\lambda=\{0.3, 0.5, 0.9\}$. Higher CTC weight gave lower WERs when 
training on low amounts (e.g. 50hr) of data. The CTC weight ($\alpha=\{0.0, 0.1, 0.5\}$) during ST training and decoding ($\beta=\{0.0, 0.1, 0.3, 0.5, 0.7, 0.9\}$) are the main hyperparameters explored in our experiments, while keeping the rest of the network architecture the same across all the ASR and ST setups. Decoding is done with beam size 10, while best $\beta$ was chosen based on performance on development set. 

All the text is tokenized using \textsc{moses} toolkit. We retain the true-case and punctuation for both ASR and ST experiments, which allowed us to use the same vocabulary of tokens for both ASR and ST models. Unigram-based segmentation method~\cite{kudo-2018-subword} from SentencePiece~\cite{kudo-richardson-2018-sentencepiece} was used to learn the sub-word vocabulary of 1000 tokens for each language. The subword segmentation algorithm was trained only on the text transcripts from ASR training data. In case of random initialization of ST models, the ST training data was used of learning the segmentation.
\subsection{Training details}
All the monolingual ASR models have $27.93$M parameters, whereas the multilingual ASR has $31.78$M. Depending on the size of training data, it took between 6 - 30 hours on a single GPU to train these models.
The ST models were initialized from pre-trained ASR models in two ways: (i) retain the CTC layer and perform joint training, (ii) discard the CTC layer to perform standard training with attention loss. In case of initialization from multilingual ASR model, the parameters of non-target language do not get updated.
All our experiments were conducted on a custom clone\footnote{\texttt{\url{https://github.com/BUTSpeechFIT/espnet/tree/main/egs2/iwslt22_low_resource/st1}}} of ESPnet2 
framework~\cite{inaguma-etal-2020-espnet}. 
\begin{figure}[!t]
  \centering
  \includegraphics[width=\linewidth]{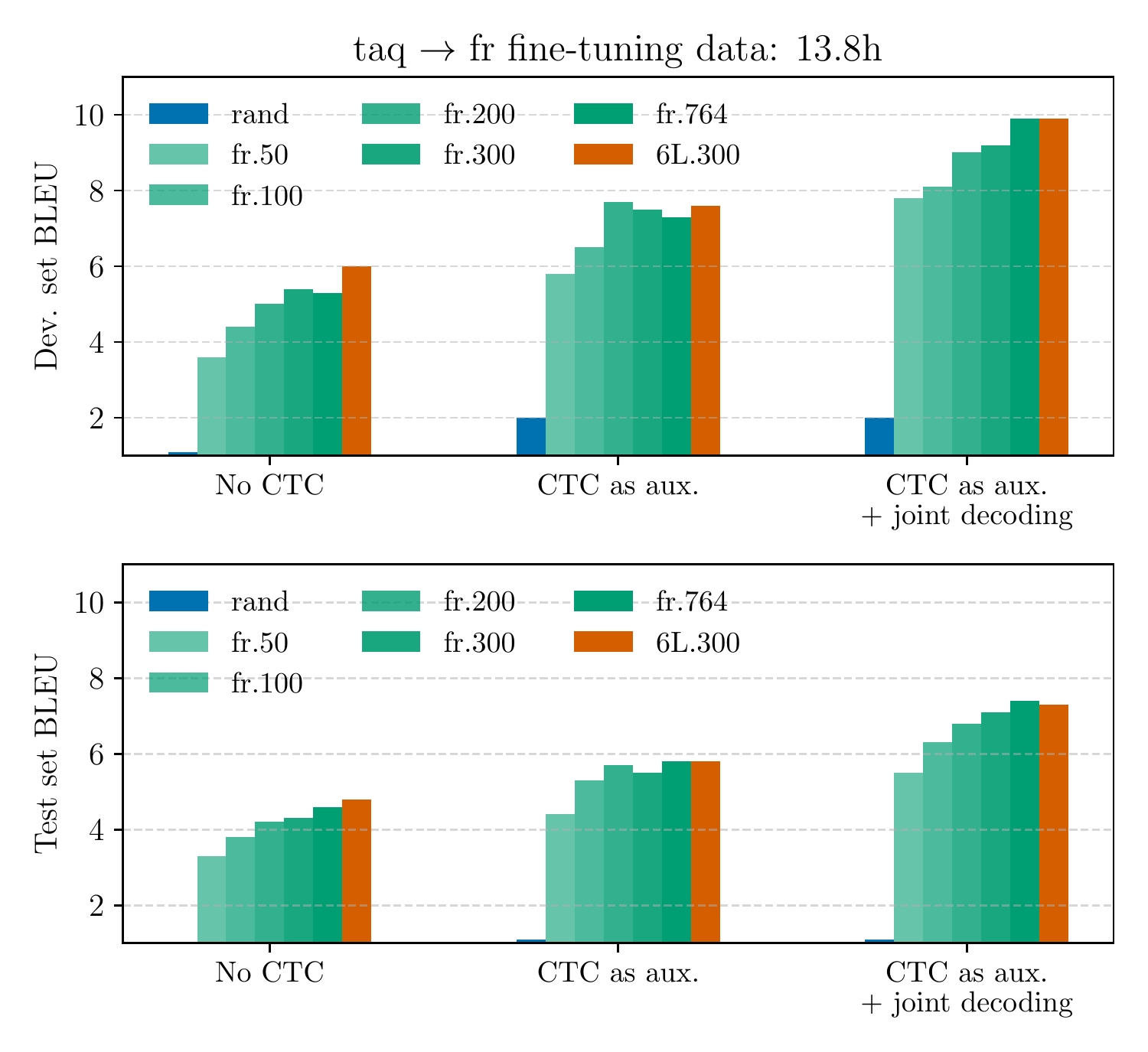}
  \caption{Performance of ST systems on taq $\rightarrow$ fr dataset, relying on various initialization, fine-tuning and decoding schemes.}
  \label{fig:taq_fr}
  \vspace{-0.25cm}
\end{figure}
\begin{figure*}[!ht]
  \centering
  \includegraphics[width=\linewidth]{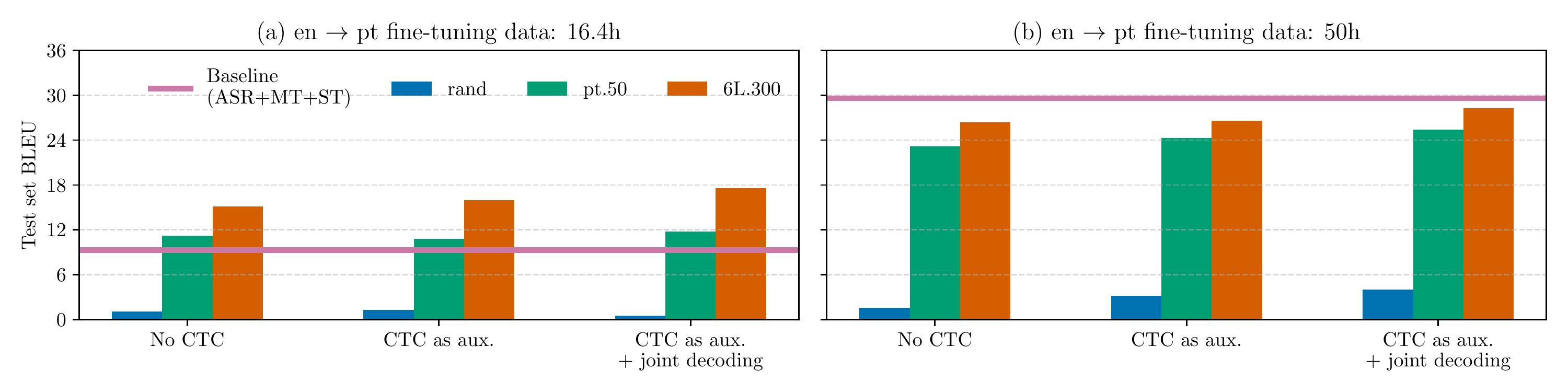}
  \caption{Effect of various initialization and amounts of ST fine-tuning data.}
  \label{fig:how2_en_pt}
  \vspace{-0.3cm}
\end{figure*}
\section{Results and discussion}
\label{sec:res}
This section presents the results of ASR and ST systems. Since we trained ASR models on true case text with punctuation, the word-error-rates (WER) would be slightly higher than if we were to train on lower case text. Hence, we report both WER and character error rate (CER) for ASR systems. The ST systems were evaluated using 4-gram BLEU with the help of sacrebleu~\cite{post-2018-call} library\footnote{nrefs:1\textpipe case:mixed\textpipe eff:no\textpipe tok:13a\textpipe smooth:exp\textpipe version:2.3.1}. We additionally report \chrf \footnote{nrefs:1\textpipe case:mixed\textpipe eff:yes\textpipe nc:6\textpipe nw:0\textpipe space:no\textpipe version:2.3.1}, an F-score based on character $n$-grams~\cite{popovic-2015-chrf}.

Table~\ref{tab:asr_res} presents the results of our ASR systems in terms of word and character error rates (WER, CER). In case of French (fr), we can see that the multilingual ASR model performs worse than the best monolingual ASR in terms of WER. The difference in CER is a bit lower. This is caused by smaller model capacity ($d_{\mathrm{model}}=256$). However, we still report results with this model, as it would be comparable to the monolingual counter-parts in terms of architecture and  parameters.

The ST models initialized from pre-trained ASR are fine-tuned in two ways (i) no CTC (ii) CTC as auxiliary objective. Once the ST model is fine-tuned, the beam-search based decoding can use either CTC score (joint decoding) or not. The Fig.~\ref{fig:taq_fr} illustrates the performance of ST system relying on various initializations, fine-tuning and decoding schemes. We can observe three things from the Fig.~\ref{fig:taq_fr}
\begin{enumerate}
  \item CTC as auxiliary objective for translation helps across various initializations. Joint CTC decoding gives further benefits.
  \item Target language ASR models (fr.50, fr.100, $\ldots$, fr.764) act as good initializations (which was also observed in prior works~\cite{BansalKLLG19}) for speech translation.
  \item The multilingual model trained on 300 hours of speech (6L 300h), which includes only 50 hours of target French data, performs better than most of the French monolingual models trained on much larger data. This suggests that even if the target-language has low-to-moderate amount of transcribed speech, one can rely on a multilingual ASR model.
\end{enumerate}
  Table~\ref{tab:taq_fr} compares our best systems (from Fig~\ref{fig:taq_fr}) with the results reported in the findings of IWSLT'22~\cite{iwslt-2022-findings}.

  With the low-resource simulation experiments (en $\rightarrow$ pt), we aim to identify saturation of benefits from pre-trained ASR, given source language transcriptions and source $\rightarrow$ target parallel data. We trained two source language (en) ASR models on 16.4 and 50 hours of transcribed speech, respectively (Table~\ref{tab:st_data_stats}). Then, we trained two en $\rightarrow$ pt MT systems on the corresponding parallel sentences (10.5k, 31.5k). We used the speech encoder from ASR and decoder from MT model to initialize an ST model, which was then fine-tuned on 16.4 and 50 hours respectively. During this fine-tuning, we use source language transcripts as targets for CTC objective function (Fig~\ref{fig:block}c). This baseline is represented by (ASR+MT+ST). Fig.~\ref{fig:how2_en_pt} shows the BLEU score on test set for all kinds of initializations. Under the low-resource setup of 16.4h, we can see that models based on target-language pre-trained ASR outperform the baseline by a decent margin. In case of mid-resource setup, i.e., with 50 hours of data, the gap reduces to 1 BLEU score. Both Fig~\ref{fig:taq_fr} and~\ref{fig:how2_en_pt} have same trends, that CTC as auxiliary objective for translation and joint decoding is beneficial.
  We also experimented with various CTC weights ($\alpha$) during training. While in most of the low-resource setups, $\alpha=0.1$ seemed to give best result. As the amount of ST fine-tuning data increased, we observed that higher CTC weight $\alpha=0.5$ yielded better results. However, a further investigation on the influence of pre-trained multilingual ASR models in high-resource setups is required.
\begin{table}[!ht]
  \centering
  \setlength{\tabcolsep}{3pt}
  \caption{Performance of ST systems on taq $\rightarrow$ fr. $^\dagger$The findings of IWSLT~\cite{iwslt-2022-findings} reports \textsc{chrF++}, however their sacrebleu signature (footnote 30) with option \texttt{nw:0} suggests that it is \textsc{chrF}, with an unknown $\beta$. Hence, the numbers cannot be compared.}
  \label{tab:taq_fr}
  \begin{tabular}{lrr|rr} \toprule
    & \multicolumn{2}{c|}{Dev.} & \multicolumn{2}{c}{Test} \\
    System     & BLEU & \chrf & BLEU & \chrf \\
    \midrule
    Wav2vec2 (taq) + ST~\cite{iwslt-2022-findings} & 8.3 & - & 5.7 & 
    31.4$^\dagger$ \\
    ASR + ST~\cite{boito-etal-2022-speech} & 6.4 & - & 5.0 & - \\
    XLS-R + mBART~\cite{iwslt-2022-findings} & - & - & 2.7 & 24.3$^\dagger$ \\
    \midrule
    Mono (fr 764h) + ST  & 9.9 & 35.2 & \textbf{7.4} & 30.9 \\
    Multi (6L 300h) + ST & 9.9 & 34.9 & \textbf{7.3} & 30.5 \\
    \bottomrule
  \end{tabular}
    \vspace{-0.1cm}
\end{table}
\section{Conclusion}
In this paper, we revisited several strategies for improving low-resource speech translation. We combined recent findings from joint-training and decoding in ASR and direct speech translation techniques and studied them with-respect-to various initializations in low-resource scenarios. Our experiments re-confirmed prior works that target-language ASR acts as good initialization for downstream speech translation. In addition, we found that pre-trained multilingual ASR is a viable alternative and performs better than the monolingual ASR in a majority of the settings. Finally, with only 300 hours of pre-training, our approaches achieved 7.3 BLEU score on low-resource Tamasheq - French dataset, outperforming prior works from IWSLT 2022.

In the future, we would like to study the effect of multilingual ST fine-tuning, as it should provide additional supervision, thus help the overall translation quality. Another important direction relates to quantifying misaligned representations when initializing modules from different modalities.
\section{Acknowledgements}
The work was supported by Horizon 2020 Marie Sk\l{}odowska-Curie grant ESPERANTO, No. 101007666, Czech National Science Foundation (GACR) project NEUREM3 No. 19-26934X, Czech Ministry of Interior project VK01020132. Computing on IT4I supercomputer was supported by the Czech Ministry of Education, Youth and Sports through the e-INFRA CZ (ID:90254). This work was inspired by insights gained from JSALT 2022, which was supported by Amazon, Microsoft and Google.

\bibliographystyle{IEEEtran}
\bibliography{mybib}

\begin{thebibliography}{10}
\providecommand{\url}[1]{#1}
\csname url@samestyle\endcsname
\providecommand{\newblock}{\relax}
\providecommand{\bibinfo}[2]{#2}
\providecommand{\BIBentrySTDinterwordspacing}{\spaceskip=0pt\relax}
\providecommand{\BIBentryALTinterwordstretchfactor}{4}
\providecommand{\BIBentryALTinterwordspacing}{\spaceskip=\fontdimen2\font plus
\BIBentryALTinterwordstretchfactor\fontdimen3\font minus
  \fontdimen4\font\relax}
\providecommand{\BIBforeignlanguage}[2]{{%
\expandafter\ifx\csname l@#1\endcsname\relax
\typeout{** WARNING: IEEEtran.bst: No hyphenation pattern has been}%
\typeout{** loaded for the language `#1'. Using the pattern for}%
\typeout{** the default language instead.}%
\else
\language=\csname l@#1\endcsname
\fi
#2}}
\providecommand{\BIBdecl}{\relax}
\BIBdecl

\bibitem{Hari:2020:Joint}
H.~K. Vydana, M.~Karafiát, K.~Zmolikova, L.~Burget, and J.~Černocký,
  ``{Jointly Trained Transformers Models for Spoken Language Translation},'' in
  \emph{IEEE International Conference on Acoustics, Speech and Signal
  Processing (ICASSP)}, 2021, pp. 7513--7517.

\bibitem{dalmia-etal-2021-searchable}
\BIBentryALTinterwordspacing
S.~Dalmia, B.~Yan, V.~Raunak, F.~Metze, and S.~Watanabe, ``{Searchable Hidden
  Intermediates for End-to-End Models of Decomposable Sequence Tasks},'' in
  \emph{Proc. of the NAACL: HLT}.\hskip 1em plus 0.5em minus 0.4em\relax
  Online: ACL, Jun. 2021, pp. 1882--1896. [Online]. Available:
  \url{https://aclanthology.org/2021.naacl-main.151}
\BIBentrySTDinterwordspacing

\bibitem{inaguma-etal-2020-espnet}
\BIBentryALTinterwordspacing
H.~Inaguma, S.~Kiyono, K.~Duh, S.~Karita, N.~Yalta, T.~Hayashi, and
  S.~Watanabe, ``{{ESP}net-{ST}: All-in-One Speech Translation Toolkit},'' in
  \emph{Proc. of the 58th Annual Meeting of the ACL: System
  Demonstrations}.\hskip 1em plus 0.5em minus 0.4em\relax Online: ACL, Jul.
  2020, pp. 302--311. [Online]. Available:
  \url{https://aclanthology.org/2020.acl-demos.34}
\BIBentrySTDinterwordspacing

\bibitem{ZhangHS22}
B.~Zhang, B.~Haddow, and R.~Sennrich, ``{Revisiting End-to-End Speech-to-Text
  Translation From Scratch},'' in \emph{International Conference on Machine
  Learning}, ser. Proc. of Machine Learning Research, K.~Chaudhuri, S.~Jegelka,
  L.~Song, C.~Szepesv{\'{a}}ri, G.~Niu, and S.~Sabato, Eds., vol. 162.\hskip
  1em plus 0.5em minus 0.4em\relax {PMLR}, July 2022, pp. 26\,193--26\,205.

\bibitem{strassel-tracey-2016-lorelei}
\BIBentryALTinterwordspacing
S.~Strassel and J.~Tracey, ``{{LORELEI} Language Packs: Data, Tools, and
  Resources for Technology Development in Low Resource Languages},'' in
  \emph{Proc. of {LREC}}.\hskip 1em plus 0.5em minus 0.4em\relax
  Portoro{\v{z}}, Slovenia: European Language Resources Association (ELRA), May
  2016, pp. 3273--3280. [Online]. Available:
  \url{https://aclanthology.org/L16-1521}
\BIBentrySTDinterwordspacing

\bibitem{BansalKLG20}
S.~Bansal, H.~Kamper, A.~Lopez, and S.~Goldwater, ``{Cross-Lingual Topic
  Prediction For Speech Using Translations},'' in \emph{Proc. of
  {ICASSP}}.\hskip 1em plus 0.5em minus 0.4em\relax {IEEE}, May 2020, pp.
  8164--8168.

\bibitem{BansalKLLG19}
S.~Bansal, H.~Kamper, K.~Livescu, A.~Lopez, and S.~Goldwater, ``{Pre-training
  on high-resource speech recognition improves low-resource speech-to-text
  translation},'' in \emph{Proc. of the NAACL:HLT}, J.~Burstein, C.~Doran, and
  T.~Solorio, Eds.\hskip 1em plus 0.5em minus 0.4em\relax ACL, June 2019, pp.
  58--68.

\bibitem{xls-r:2022}
\BIBentryALTinterwordspacing
A.~Babu, C.~Wang, A.~Tjandra, K.~Lakhotia, Q.~Xu, N.~Goyal, K.~Singh, P.~von
  Platen, Y.~Saraf, J.~Pino, A.~Baevski, A.~Conneau, and M.~Auli, ``{{XLS-R:}
  Self-supervised Cross-lingual Speech Representation Learning at Scale},'' in
  \emph{Proc. of Interspeech}, H.~Ko and J.~H.~L. Hansen, Eds.\hskip 1em plus
  0.5em minus 0.4em\relax {ISCA}, September 2022, pp. 2278--2282. [Online].
  Available: \url{https://doi.org/10.21437/Interspeech.2022-143}
\BIBentrySTDinterwordspacing

\bibitem{StoianBG20}
M.~C. Stoian, S.~Bansal, and S.~Goldwater, ``{Analyzing {ASR} Pretraining for
  Low-Resource Speech-to-Text Translation},'' in \emph{Proc. of
  {ICASSP}}.\hskip 1em plus 0.5em minus 0.4em\relax {IEEE}, May 2020, pp.
  7909--7913.

\bibitem{mbart:2020}
\BIBentryALTinterwordspacing
Y.~Liu, J.~Gu, N.~Goyal, X.~Li, S.~Edunov, M.~Ghazvininejad, M.~Lewis, and
  L.~Zettlemoyer, ``{Multilingual Denoising Pre-training for Neural Machine
  Translation},'' \emph{Transactions of the ACL}, vol.~8, pp. 726--742, 2020.
  [Online]. Available: \url{https://aclanthology.org/2020.tacl-1.47}
\BIBentrySTDinterwordspacing

\bibitem{chuang-etal-2021-investigating}
\BIBentryALTinterwordspacing
S.-P. Chuang, Y.-S. Chuang, C.-C. Chang, and H.-y. Lee, ``{Investigating the
  Reordering Capability in {CTC}-based Non-Autoregressive End-to-End Speech
  Translation},'' in \emph{Findings of the ACL: ACL-IJCNLP 2021}.\hskip 1em
  plus 0.5em minus 0.4em\relax Online: ACL, Aug. 2021, pp. 1068--1077.
  [Online]. Available: \url{https://aclanthology.org/2021.findings-acl.92}
\BIBentrySTDinterwordspacing

\bibitem{bentivogli-etal-2021-cascade}
\BIBentryALTinterwordspacing
L.~Bentivogli, M.~Cettolo, M.~Gaido, A.~Karakanta, A.~Martinelli, M.~Negri, and
  M.~Turchi, ``{Cascade versus Direct Speech Translation: Do the Differences
  Still Make a Difference?}'' in \emph{Proc. of the 59th Annual Meeting of the
  ACL and the 11th IJCNLP}.\hskip 1em plus 0.5em minus 0.4em\relax Online: ACL,
  Aug. 2021, pp. 2873--2887. [Online]. Available:
  \url{https://aclanthology.org/2021.acl-long.224}
\BIBentrySTDinterwordspacing

\bibitem{joint_vs_cascade_2022}
\BIBentryALTinterwordspacing
V.~A.~K. Tran, D.~Thulke, Y.~Gao, C.~Herold, and H.~Ney, ``{Does Joint Training
  Really Help Cascaded Speech Translation?}'' in \emph{Proc. of Conference on
  EMNLP}.\hskip 1em plus 0.5em minus 0.4em\relax Abu Dhabi, United Arab
  Emirates: ACL, Dec. 2022, pp. 4480--4487. [Online]. Available:
  \url{https://aclanthology.org/2022.emnlp-main.297}
\BIBentrySTDinterwordspacing

\bibitem{yan:2023:eacl}
\BIBentryALTinterwordspacing
B.~Yan, S.~Dalmia, Y.~Higuchi, G.~Neubig, F.~Metze, A.~W. Black, and
  S.~Watanabe, ``{CTC} alignments improve autoregressive translation,'' in
  \emph{Proceedings of the 17th Conference of the EACL}.\hskip 1em plus 0.5em
  minus 0.4em\relax Dubrovnik, Croatia: ACL, May 2023, pp. 1623--1639.
  [Online]. Available: \url{https://aclanthology.org/2023.eacl-main.119}
\BIBentrySTDinterwordspacing

\bibitem{Graves:2006:CTC}
\BIBentryALTinterwordspacing
A.~Graves, S.~Fern\'{a}ndez, F.~Gomez, and J.~Schmidhuber, ``{Connectionist
  Temporal Classification: Labelling Unsegmented Sequence Data with Recurrent
  Neural Networks},'' in \emph{Proc. of the 23rd ICML}, ser. ICML '06.\hskip
  1em plus 0.5em minus 0.4em\relax New York, NY, USA: Association for Computing
  Machinery, 2006, p. 369–376. [Online]. Available:
  \url{https://doi.org/10.1145/1143844.1143891}
\BIBentrySTDinterwordspacing

\bibitem{yan-etal-2022-cmus}
\BIBentryALTinterwordspacing
B.~Yan, P.~Fernandes, S.~Dalmia, J.~Shi, Y.~Peng, D.~Berrebbi, X.~Wang,
  G.~Neubig, and S.~Watanabe, ``{{CMU}{'}s {IWSLT} 2022 Dialect Speech
  Translation System},'' in \emph{Proc. of the 19th {IWSLT}}.\hskip 1em plus
  0.5em minus 0.4em\relax Dublin, Ireland (in-person and online): ACL, May
  2022, pp. 298--307. [Online]. Available:
  \url{https://aclanthology.org/2022.iwslt-1.27}
\BIBentrySTDinterwordspacing

\bibitem{iwslt-2022-findings}
\BIBentryALTinterwordspacing
A.~Anastasopoulos, L.~Barrault, L.~Bentivogli, M.~Zanon~Boito, O.~Bojar,
  R.~Cattoni, A.~Currey, G.~Dinu, K.~Duh, M.~Elbayad, C.~Emmanuel,
  Y.~Est{\`e}ve, M.~Federico, C.~Federmann, S.~Gahbiche, H.~Gong,
  R.~Grundkiewicz, B.~Haddow, B.~Hsu, D.~Javorsk{\'y}, V.~Kloudov{\'a},
  S.~Lakew, X.~Ma, P.~Mathur, P.~McNamee, K.~Murray, M.~N{\v{a}}dejde,
  S.~Nakamura, M.~Negri, J.~Niehues, X.~Niu, J.~Ortega, J.~Pino, E.~Salesky,
  J.~Shi, M.~Sperber, S.~St{\"u}ker, K.~Sudoh, M.~Turchi, Y.~Virkar, A.~Waibel,
  C.~Wang, and S.~Watanabe, ``{Findings of the {IWSLT} 2022 Evaluation
  Campaign},'' in \emph{Proc. of the 19th {IWSLT}}.\hskip 1em plus 0.5em minus
  0.4em\relax Dublin, Ireland (in-person and online): ACL, May 2022, pp.
  98--157. [Online]. Available: \url{https://aclanthology.org/2022.iwslt-1.10}
\BIBentrySTDinterwordspacing

\bibitem{boito-etal-2022-speech}
\BIBentryALTinterwordspacing
M.~Zanon~Boito, F.~Bougares, F.~Barbier, S.~Gahbiche, L.~Barrault, M.~Rouvier,
  and Y.~Est{\`e}ve, ``{Speech Resources in the {T}amasheq Language},'' in
  \emph{Proc. of {LREC}}.\hskip 1em plus 0.5em minus 0.4em\relax Marseille,
  France: European Language Resources ACLociation, Jun. 2022, pp. 2066--2071.
  [Online]. Available: \url{https://aclanthology.org/2022.lrec-1.222}
\BIBentrySTDinterwordspacing

\bibitem{Vaswani:NIPS2017}
A.~Vaswani, N.~Shazeer, N.~Parmar, J.~Uszkoreit, L.~Jones, A.~N. Gomez, L.~u.
  Kaiser, and I.~Polosukhin, ``Attention is all you need,'' in \emph{Advances
  in Neural Information Processing Systems}, vol.~30.\hskip 1em plus 0.5em
  minus 0.4em\relax Curran Associates, Inc., 2017.

\bibitem{karita19_interspeech}
S.~Karita, N.~E.~Y. Soplin, S.~Watanabe, M.~Delcroix, A.~Ogawa, and
  T.~Nakatani, ``{Improving Transformer-Based End-to-End Speech Recognition
  with Connectionist Temporal Classification and Language Model Integration},''
  in \emph{Proc. of Interspeech}, 2019, pp. 1408--1412.

\bibitem{how2:2018}
\BIBentryALTinterwordspacing
R.~Sanabria, O.~Caglayan, S.~Palaskar, D.~Elliott, L.~Barrault, L.~Specia, and
  F.~Metze, ``{{How2:} A Large-scale Dataset For Multimodal Language
  Understanding},'' in \emph{Proc. of the Workshop on Visually Grounded
  Interaction and Language (ViGIL)}.\hskip 1em plus 0.5em minus 0.4em\relax
  NeurIPS, 2018. [Online]. Available: \url{http://arxiv.org/abs/1811.00347}
\BIBentrySTDinterwordspacing

\bibitem{kudo-2018-subword}
\BIBentryALTinterwordspacing
T.~Kudo, ``Subword regularization: Improving neural network translation models
  with multiple subword candidates,'' in \emph{Proc. of the 56th Annual Meeting
  of the ACL}.\hskip 1em plus 0.5em minus 0.4em\relax Melbourne, Australia:
  ACL, Jul. 2018, pp. 66--75. [Online]. Available:
  \url{https://aclanthology.org/P18-1007}
\BIBentrySTDinterwordspacing

\bibitem{kudo-richardson-2018-sentencepiece}
\BIBentryALTinterwordspacing
T.~Kudo and J.~Richardson, ``{{S}entence{P}iece: A simple and language
  independent subword tokenizer and detokenizer for Neural Text Processing},''
  in \emph{Proc. of the 2018 Conference on EMNLP: System Demonstrations}.\hskip
  1em plus 0.5em minus 0.4em\relax Brussels, Belgium: ACL, Nov. 2018, pp.
  66--71. [Online]. Available: \url{https://aclanthology.org/D18-2012}
\BIBentrySTDinterwordspacing

\bibitem{post-2018-call}
\BIBentryALTinterwordspacing
M.~Post, ``{A Call for Clarity in Reporting {BLEU} Scores},'' in \emph{Proc. of
  the Third Conference on Machine Translation: Research Papers}.\hskip 1em plus
  0.5em minus 0.4em\relax Brussels, Belgium: ACL, Oct. 2018, pp. 186--191.
  [Online]. Available: \url{https://aclanthology.org/W18-6319}
\BIBentrySTDinterwordspacing

\bibitem{popovic-2015-chrf}
\BIBentryALTinterwordspacing
M.~Popovi{\'c}, ``{chr{F}: character n-gram {F}-score for automatic {MT}
  evaluation},'' in \emph{Proc. of the Tenth Workshop on Statistical Machine
  Translation}.\hskip 1em plus 0.5em minus 0.4em\relax Lisbon, Portugal: ACL,
  Sep. 2015, pp. 392--395. [Online]. Available:
  \url{https://aclanthology.org/W15-3049}
\BIBentrySTDinterwordspacing

\end{thebibliography}

\end{document}